%% file: arxiv.tex
\pdfoutput=1

\documentclass[11pt]{article}
\usepackage{authblk}
\usepackage[]{emnlp2021}

\usepackage{times}
\usepackage{latexsym}
\usepackage{booktabs}
\usepackage{tabularx}
\usepackage{tikz}
\usepackage{pgfplots}
\usepackage{pgfplotstable}
\usepackage{enumitem}

\usepackage{soul}
\usepackage{pifont}
\usepackage{xcolor,colortbl}

\usepackage{cleveref}
\usepackage[]{mdframed}
\usepackage{framed}
\newcommand{\cmark}{\ding{51}}%
\newcommand{\xmark}{\ding{55}}%

\usepackage{url}
\usepackage{graphicx}
\usepackage{hyperref}
\hypersetup{
    colorlinks=true,
    linkcolor=blue,
    filecolor=magenta,      
    urlcolor=purple,
    pdftitle={Overleaf Example},
    pdfpagemode=FullScreen,
    }

\usepackage[T1]{fontenc}

\usepackage[utf8]{inputenc}

\usepackage{microtype}

\makeatletter 
\renewcommand\AB@affilsepx{\quad\protect\Affilfont}
\makeatother

\title{Aligning Language Models to User Opinions}

\author[$1,2$]{\textbf{EunJeong Hwang}}
\author[$3$]{\textbf{Bodhisattwa Prasad Majumder}*}
\author[$4$]{\textbf{Niket Tandon}*}
\affil[$1$]{University of British Columbia}
\affil[$2$]{Vector Institute for AI}
\affil[$3$]{UC San Diego}
\affil[$4$]{Allen Institute of AI \protect \\ \texttt{ejhwang@cs.ubc.ca, bodhisattwa@ucsd.edu, nikett@allenai.org} \protect \\ \vspace{1ex}
\small *contributes equally}

\makeatletter
\renewcommand\outauthor{
    \begin{tabular}[t]{>{\centering}p{14cm}} 
    \bf\@author
    \end{tabular}}
\makeatother

\input{macros}

\begin{document}
\maketitle
\begin{abstract}
An important aspect of developing LLMs that interact with humans is to align models' behavior to their users. It is possible to prompt an LLM into behaving as a certain persona, especially a user group or ideological persona the model captured during its pertaining stage. But, how to best align an LLM with a specific user and not a demographic or ideological group remains an open question. Mining public opinion surveys (by PEW research), we find that the opinions of a user and their demographics and ideologies are not mutual predictors. We use this insight to align LLMs by modeling both user opinions as well as user demographics and ideology, achieving up to 7 points accuracy gains in predicting public opinions from survey questions across a broad set of topics\footnote{Project page: \url{https://github.com/eujhwang/personalized-llms}}. In addition to the typical approach of prompting LLMs with demographics and ideology, we discover that utilizing the most relevant past opinions from individual users enables the model to predict user opinions more accurately.
\end{abstract}

\section{Introduction}

Personality is a defining feature of human beings, shaped by a complex interplay of demographic characteristics, moral principles, and social experiences \cite{weil1957ecrits, mclellan1989simone}. In turn, a person's personality has a significant influence on their ability to make decisions \cite{lauriola2001personality, busic2017role}. Owing to the wide-scale adaptation of the large language models (LLMs) for assisting individuals in their decision-making process \cite{Jiang2021DelphiTM, gao2023chat}, it becomes increasingly critical to ensure that these models are aligned with the unique personalities of their users. 

With lower barriers to entry, several recent works focused on prompting LLMs with persona or role-based prompts such as \texttt{Pretend you are a Democrat} \cite{deshpande2023toxicity, opinionqa}. However, the extent to which these approaches align language models with users remains unclear due to the subjective nature of defining user personas. Users have nuanced opinions that can change over time and vary depending on context. While alignment with normalized user groups like religion or political inclination may be easier, LLMs continue to struggle to align with individual users or the long tail of user groups. Additionally, LLMs tend to form opinions based on their pretraining data, as well as feedback collected from crowd workers and model designers. As a result, they exhibit low steerability, even with user groups that have major representation \cite{opinionqa}.

Aligning LLMs to individual and long-tail opinions has received less attention, while mostly focusing on aligning to user groups. In our analysis over PEW surveys, we found that people can share all of their demographic traits but still exhibit a large variance in their opinions, rendering the current group-based LLM alignment insufficient. This paper investigates the relationship between demographic traits and individual opinions in LLM alignment. Specifically, we seek to answer the following research question:
\begin{quote}
    What do we need to align an LLM to a user: demographic traits, fine-grained opinions, or both?
\end{quote}

\begin{figure*}[t!]
    \centering
    \includegraphics[width=0.92\linewidth]{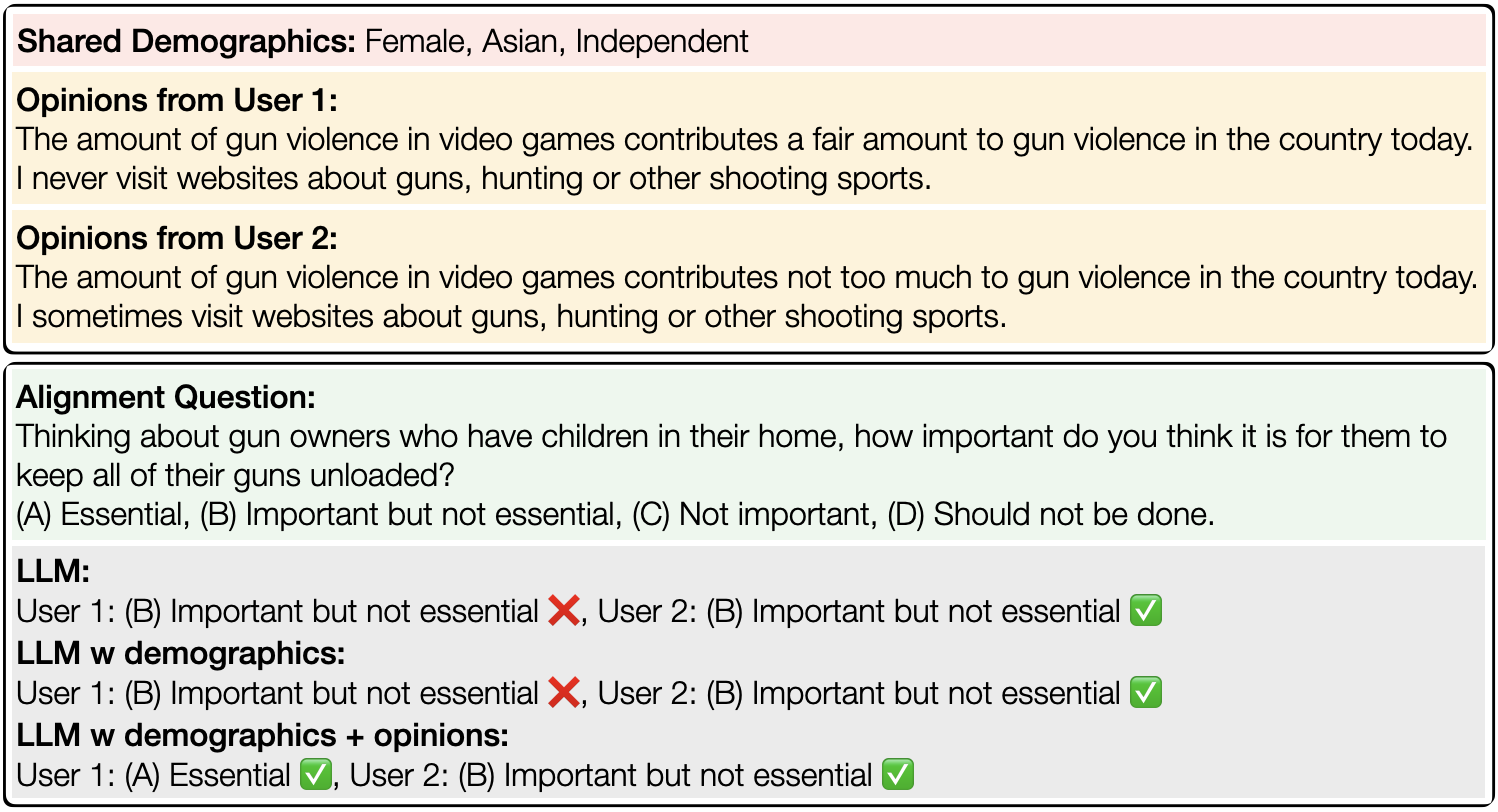}
    \caption{An illustrative example that shows opinions can vary even when two individuals have the exact same demographic traits.}
    \label{fig:example}
\end{figure*}

The majority of the past work in NLP literature focused on aligning LLMs with normalized user groups \cite{opinionqa, majumder-etal-2019-generating, salemi2023lamp}. In social science studies, however, it has been shown that all users are unique even if they belong to the same broader user group, and normalizing user groups is not a true representative of a user's opinion \cite{chu2023language, kim2023aiaugmented}. Inspired by these social science studies, we apply the insights to an empirical setting where we try to model individuals' opinions based on their various persona information such as demographic traits, ideological inclinations, and past opinions.

In this paper, we give a thorough analysis of public survey responses in the OpinionQA dataset \cite{opinionqa} with respect to their demographics, ideology, and implicit opinions and present comprehensive experimental results using the GPT3 model with various combinations of inputs (i.e., demographic, ideology, user's past opinions). Through our dataset analysis, we found that users' opinions and demographics do not necessarily correlate with each other. Our experimental results show incorporating both user opinions, demographics, and ideology, results in significant gains of up to 7 points in QA accuracy for certain topics, and utilizing the most relevant past opinions helps the model to pinpoint the more accurate answers for the users.

\section{Related work}

\paragraph{Personalization}
Past works that focused on modeling individual users were from the pre-LLMs era and mainly hail from the recommender systems literature \cite{gao2023chatrec, he2017neural, li2021selfsupervised, majumder-etal-2019-generating}. However, these systems were trained on domain-specific annotated datasets or using latent information about the users (e.g., modeling users based on their previously written reviews which generally contain sparse information about the user). The LLMs we use today have seen less content from the long tail of user groups during their pre-training phase, and there has been a lack of large-scale datasets of individual opinions until recently \cite{opinionqa}. Thus it remains an open problem whether LLMs can be aligned effectively with individual user persona and how different user information (e.g., demographic traits vs.~past opinions) influences how well an LLM can model individual's opinions. For a comprehensive comparison among all previous work, see \Cref{tab:related_work_short_summary}. 

\begin{table*}[!t]
\small
    \centering
    \begin{tabular}{p{0.4\textwidth}p{0.18\textwidth}p{0.14\textwidth}p{0.16\textwidth}}
    \toprule
  & \bf User-profile \newline explicitly observed & \bf Modeling \newline individuals &  \bf Requires\newline no training  \\ \midrule
  \textbf{Personalized generation}: OpinionQA \cite{opinionqa}, RecipeGen \cite{majumder-etal-2019-generating}, LAMP \cite{salemi2023lamp}
  & \xmark & \xmark~or \cmark \newline (mostly group)  & \xmark~or \cmark 
  \\ \midrule
  \textbf{Recommender Systems}: ChatRec \cite{gao2023chatrec}, Collaborative Filtering \cite{he2017neural}, BotPlay \cite{li2021selfsupervised} 
  & \xmark~or \cmark \newline (mostly latent) & \cmark & \xmark~or \cmark \newline (mostly supervised)
  \\ \midrule
  \textbf{Ours}
  & \cmark & \cmark~(+ group) & \cmark
  \\ \bottomrule
    \end{tabular}
    \caption{Placement of our work w.r.to related work}
    \label{tab:related_work_short_summary}
\end{table*}

\paragraph{Role of demographics and ideology}
There have been several studies investigating the correlation between ideological attitudes and psychological traits \cite{ideological-attitudes, Crockett2004, CHAN2021}. \citet{Crockett2004} analyzed the role of political ideology in consumer behavior and found that normative political ideology is central to understanding shopping as a manifestation of social and political connections. \citet{CHAN2021} found that the cognitive decision-making strategies of individuals reflected their ideological attitudes. Differently in our work, we show that ideology is not the only important factor in predicting the user's opinion using an LLM.

\paragraph{LLMs with retrieval-based approach}
Extensive prior work has used retrievals from a text corpus to aid QA \cite{madaan-etal-2022-memory, pan-etal-2019-improving-question}, or retrievals
of prior QA pairs for nearest-neighbor QA \cite{Khandelwal2020Generalization}. \citet{madaan-etal-2022-memory} uses a memory of user opinions to retrieve past relevant data points for the prompt. \citet{Khandelwal2020Generalization} extended a pre-trained language model (LM) with a k-nearest neighbors model and showed the effectiveness of the nearest neighbor search for language modeling. Our work builds upon those ideas. Differently from work on LLMs and user group level personalization, we show that LLMs can be tuned for individual users with their opinions.

\begin{table*}[!t]
    \centering
    \small
    \begin{tabular}{p{0.2\textwidth}p{0.04\textwidth}p{0.06\textwidth}p{0.05\textwidth}p{0.06\textwidth}p{0.06\textwidth}p{0.08\textwidth}p{0.05\textwidth}p{0.05\textwidth}}
    \toprule
  \bf & Guns & Auto & Gender & Sex. harass. & Biomed-food & Gender & 2050 US & Trust-Science \\ \midrule 
  Similar op. user pair & \cellcolor{lightgray} 45 & 13 & 30 & 12 & 11& 37 & 23 & 21  \\
Similar op. \& ideol. & \cellcolor{lightgray} 19 & 18 & 21 & 30 & 19 & 24 & 20 & 20  \\
  Similar op. \& diff. ideol. & \cellcolor{lightgray} 81 & 82 & 79 & 70 & 81 & 76 & 80 & 80 \\ \midrule
  & Race & Misinfo. & Privacy & Family & Econ.\newline Inequal. & Global \newline Attitudes & Politics &  \\ \midrule
  Similar op. user pair & 12 & 29 & 21 & \cellcolor{lightgray}43 & 25 & 24 & 16 & \\
  Similar op. \& ideol. & 30 & 20 & 17 & \cellcolor{lightgray}19 & 25 & 33 & 40 & \\
  Similar op. \& diff. ideol. & 70 & 80 & 83 & \cellcolor{lightgray}81 & 75 & 67 & 60 & \\
  \bottomrule
\end{tabular}
%
    \caption{Percentage of user pairs sharing similar opinions (similar op. user pair) and the percentages of similar ideologies (similar op. \& ideol.) and different ideologies (similar op. \& diff. ideol.) within user pairs sharing similar opinions. (Auto: Automation, Sex. harass.: Sexual harassment, Biomed-food: Biomedical food, Misinfo.: Misinformation, Econ. Inequal.: Economic Inequality, Politics: Political Views)}
    \label{tab:same_op_diff_demo}
\end{table*}

\section{What makes a persona?}

We present a study on various components that makes a personality (in short, persona) of a user. We use the OpinionQA dataset, which contains 15 topics, and each topic contains an average of 100 questions and 5340 users \cite{opinionqa}.

\subsection{Demographics}
The dataset records eight demographic information of a user: region, sex, age, education, race, citizen, marital status, and income. These are the markers of social experience that a user is most likely to go through. For example, the social experience can be determined by the region a user belongs, or their age determines whom they socialize with on a regular basis. However, this runs with the risk of stereotyping (i.e., an old individual is less likely to mix with younger people or they are conservative in thinking). We later show that demographic information is not enough to model an individual. 

\subsection{Ideology}
Ideology is formed by an individual understanding of politics and economics. In our dataset, we have each subject's political affiliation and inclinations toward well-known political ideologies (e.g., conservative, liberal). We use this information as an individual's ideology.

\subsection{Opinions}
OpinionQA uses a well-established method of capturing human opinions from public opinion surveys. In these surveys, subjects are asked to answer subjective questions that reflect their unique opinions and what makes them different from other individuals. \Cref{fig:example} shows an example of opinions that a user provided during a survey.

\subsection{Deriving insights from public surveys}
\label{sec:ds-analysis}
We derive insights from the OpinionQA dataset, where
we analyze the degree of agreement in user's opinions where they same demographics and how this agreement varies across topics. This statistical analysis generates useful insights that we later use for our modeling approach. We also look for similar (dis)agreements in opinions when users have the same ideologies.

\begin{figure}[!h]
    \includegraphics[width=0.47\textwidth]{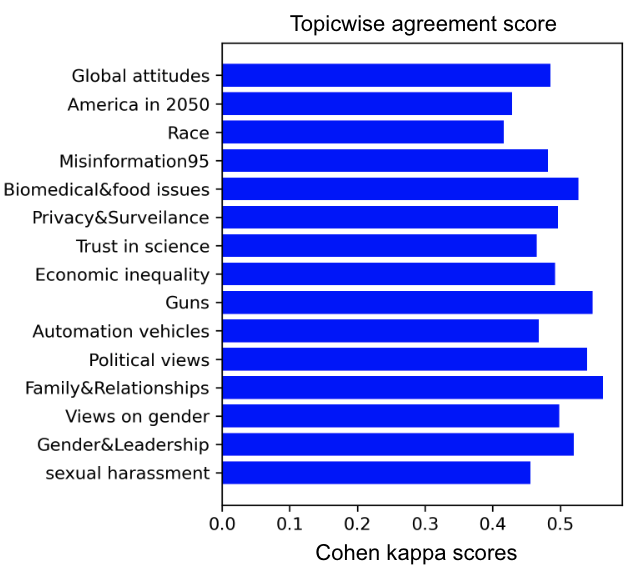}
    \caption{Topic-wise agreement score; x-axis: agreement score, y-axis: topic. This graph shows that users with similar demographics/ ideology can have different opinions (cohen kappa scores of around 0.4 show not some but not substantial correlation in opinions)}
    \label{fig:topicwise-agreement-score}
\end{figure}

\paragraph{Opinions differ despite the same demographics} We first take all pairs of users sharing the same demographics and compare their opinions. To calculate the agreement score between users, we utilize Cohen's kappa coefficient \cite{cohen-kappa}, which ranges from $-1$ to $1$. Even though two users share the same demographics, agreement scores on the implicit opinions are gathered around 0.5 (\Cref{fig:topicwise-agreement-score}). This shows that solely relying on demographic information is not enough to personalize the model, and users' implicit opinions can play a critical role in personalization.

\paragraph{Opinions differ across topics} In Figure \ref{fig:topicwise-agreement-score}, we also show the topic-wise agreement scores. On certain topics, including Family \& Relationships and Guns, users exhibit relatively higher agreement scores. On the other hand, for some topics, including Race and America in 2050, users have lower agreement scores, indicating that certain topics may have larger variability in terms of user opinions. We later analyze if this variability appears in a model's predictive performance when it is used to predict user opinions across different topics.

\paragraph{Opinions differ despite same ideology}
To analyze the correlation between user opinions and their ideology, we extract user pairs that two users who answered at least more than 10 common questions and compare their opinions and political ideologies. Table \ref{tab:same_op_diff_demo} shows the percentage of user pairs sharing similar opinions, where 70\% of opinions are matched between two users, and the percentages of the same ideologies and different ideologies within those user pairs. We observe that even though the users have similar opinions, around 80\% of the user pairs have different ideologies. In contrast, we observe the percentage of sharing similar opinions among the users having similar ideologies is relatively higher than the percentage of sharing similar ideologies among the users having similar opinions in Appendix \ref{tab:same_demo_diff_op}. This implies that while having similar opinions does not necessarily imply shared ideologies among users, the presence of similar ideologies may suggest that users are more likely to have similar opinions. We particularly notice this phenomenon on the Guns and Family topics, as \colorbox{lightgray}{highlighted} in Table \ref{tab:same_op_diff_demo}. While the percentage of user pairs with shared opinions is higher compared to other topics, the percentage of user pairs with differing ideologies within these pairs is notably higher than the percentage of user pairs with similar ideologies.

Based on the insights derived above, we incorporate them in our modeling approaches and analyze if these translate to the predictive performance of a model when used to predict user opinions as collected from the surveys.

\section{Aligning LLMs with persona}
In this section, we detail our task, possible modeling approaches, and evaluation protocols in \Cref{sec:ex-setup} and discuss how to select the most relevant past opinions of a user in \Cref{sec:llm-all-inputs}. 

\subsection{Setup}
\label{sec:ex-setup}
\paragraph{Task} We use LLMs to model a user; however, to concretely measure the performance, we use a simple question-answering (QA) setup. For our QA task, we use existing questions from the surveys and try to predict the choice from multiple-choice originally given to the subjects. We use a prompting-based zero-shot approach to perform the multiple-choice QA. We use \texttt{text-davinci-003} as the LLM. 

\paragraph{Modeling Approaches} We sample 100 users per topic. 20\% of implicit questions belonging to the specific user are used as the user's implicit persona, and the rest are used to test the model's personalization ability. We have the following variants of our model where the model is gradually exposed to different levels of user information: demographic, ideological information, and user past opinions. Here is a rough sketch of what a prompt would contain for each modeling variation:

\begin{enumerate}
    \item  \underline{no persona}: this is a case where default LLM opinion is evaluated w.r.to the individual's opinion \cite{opinionqa}.
    \item \underline{ideology}: here, we observe if ideological inclinations from the user help the model to align better to them \cite{opinionqa}.
    \item \underline{ideology + demographics}: here, we observe if both demographic information and ideological inclinations from the user help the model to align better with them \cite{opinionqa}.
    \item \underline{ideology + opinions}: we combine ideological inclinations and opinions and measure if these help the model to align better with an individual.
    \item \underline{demographic + ideology + opinions}: here, we observe when we combine all possible personal information, i.e., demographic, ideology, and opinions, and measure if these help the model to align better with an individual. See \Cref{fig:prompt:imexp} for the complete prompt.
\end{enumerate}

\paragraph{Evaluation Metric} For evaluation, we utilize two types of accuracy measures, \emph{overall accuracy} and \emph{collapsed accuracy}. For overall accuracy, we simply calculate the accuracy of the precited answer choice with respect to the gold answer choice from the dataset. We also present collapsed accuracy because most answer choices in the opinion QA dataset have around 3 to 4 classes. In cases where there are more than 4 classes, it is possible to further group the classes into superclasses without losing substantial finer information. For example, the following answer choices: \texttt{[Very likely, Somewhat likely, Not too likely, Not at all likely]}, can be grouped into \texttt{[Likely, Unlikely]}. We consolidate such answer choices into two classes, referred to as \emph{collapsed accuracy}, and present the results accordingly.

\input{prompts/prompt-demo-ideo-opinion}

\subsection{LLM as a person with ideologies, demographic, and opinions}
\label{sec:llm-all-inputs}

\begin{table*}[t!]
\small
    \centering
\begin{tabular}{@{}lcc@{}}
\toprule
 \bf Model & \bf Exact match & \bf Collapsed match  \\
\midrule
no persona & 0.43$\pm$0.01 & 0.62$\pm$0.01 \\
demographic + ideology & 0.47$\pm$0.01 & 0.65$\pm$0.01 \\
demographic + ideology + all opinions & 0.51$\pm$0.01 & 0.69$\pm$0.01 \\ \midrule
ideology + top-8 opinions & 0.53$\pm$0.01 & 0.69$\pm$0.01 \\
demographic + top-8 opinions & 0.53$\pm$0.01 & 0.69$\pm$0.01 \\
demographic + ideology + top-3 opinions & 0.53$\pm$0.01 & 0.69$\pm$0.01 \\ \midrule
top-3 opinions & 0.51$\pm$0.01 & 0.67$\pm$0.01 \\
top-8 opinions & 0.52$\pm$0.01 & 0.68$\pm$0.01 \\
demographic + ideology + top-8 opinions & \textbf{0.54$\pm$0.01} & \textbf{0.70$\pm$0.01} \\
\bottomrule
\end{tabular}
%
\caption{Overall QA accuracy. For statistical significance, we computed Wilson score intervals for $\alpha$= 99\%
}
\label{tab:qa-accuracy}
\end{table*}

Our main goal is to use different components of a user's persona (demographics, ideology, opinions) to align an LLM with an individual.
Specifically, by having two experiments, one with past opinions+ideology and the other with past opinions+ideology+demographics, we aim to analyze the role of demographics when predicting user responses. In addition, we hypothesize that giving users' past opinions may offer useful insights into their perspective (followed from \Cref{tab:same_op_diff_demo}), and LLM can benefit from that information when predicting the future answer for the specific user. When adding the user's past opinions, we compare the model with all opinions (maximum 16) to the model with top-$k$ opinions ($k$ is a hyperparameter $\in {3, 5, 8}$). The top-$k$ opinions are obtained by comparing the embedding similarity between the user's previous opinions and the question at hand, where we employ \texttt{text-embedding-ada-002} to obtain the embeddings. We hypothesize that all opinions may incorporate some unrelated viewpoints to answer the question, and hence offering more pertinent opinions would enhance the model's ability to accurately anticipate its future response for the user. \Cref{fig:prompt:imexp} shows a complete prompt where use all available past information of individuals to predict their future opinions. Other modeling approaches noted in \Cref{sec:ex-setup} have ablated versions of this prompt according to their descriptions given (see Appendix \ref{appendix:prompt}).

\section{Results and Analysis}
\label{sec:analysis}
Here, we first analyze our model variants (\Cref{sec:model-analysis}) to validate hypotheses that we gather from analyzing the dataset (in \Cref{sec:ds-analysis}). We also provide our model's performance when we use a similar modeling setup to predict group-level opinions in \Cref{sec:llm-majority,sec:llm-ideol}. 

\subsection{LLM for an individual} 
\label{sec:model-analysis}
Here we discuss the results of using an LLM to model an individual in the light of the evaluation metrics described in \Cref{sec:model-analysis}.

\paragraph{Exact match vs. Collapsed match} 
The accuracy with the exact match and with the collapsed match in \Cref{tab:qa-accuracy} and \Cref{tab:topicwise-accuracy} shows a similar trend for the performance of our model variants. Especially with topic-wise collapsed accuracy in \Cref{tab:topicwise-accuracy}, the model variant that incorporates demographic information and user's past opinions outperforms in most topics, exhibiting a more substantial margin compared to the variant that solely incorporates demographic information. This suggests that leveraging implicit opinions enables the model to align with the correct range of answer choices, even though it does not precisely predict the exact same answer as the user's choice.

\paragraph{Overall Accuracy} 
\Cref{tab:qa-accuracy} presents overall QA accuracy with exact match and collapsed match for answer choices. Adding demographic and ideology information outperforms the model without any persona, indicating that some questions might be highly correlated with the user's demographics, and LLM is able to make a guess with the demographic information. Incorporating the user's previous opinions, up to 16 in total, along with demographic information, substantially enhances the performance in both overall and collapsed accuracy. This implies that users' past opinions are indeed important to make correct predictions. 

Interestingly, utilizing the top-$k$ most relevant previous opinions does not yield a significant increase in collapsed accuracy. However, it does improve the exact match accuracy by up to 3 points when using both demographics and ideology along with the user's previous opinions. This implies that having top-k most relevant past opinions can help the model pinpoint more accurate answers, and providing the user's past opinions is already pushing the model to be in the correct range of the answer choices. We noticed that utilizing the top-3 opinions yields similar performance to using the top-8 opinions, indicating that a few of the most relevant opinions carry the most performance improvement of the model. Moreover, simply using the top 3 most relevant opinions performs on par with the model with user demographic, ideology, and user's past 16 random opinions. This confirms again that utilizing the most relevant opinions as feedback is essential to get personalized answers from LLM. Lastly, providing additional demographic information with ideology slightly improves the model performance, implying that the demographic information may contribute valuable insights to the model to a certain degree. 

\begin{table*}[t!]
\small
\centering
\begin{tabular}{lccc
}
\toprule
 & \multicolumn{3}{c}{\bf Accuracy with exact match}  
 \\
 & no-persona & demo. + ideo. & demo. + ideo.+ top8 op. 
 \\
\midrule
Guns & 0.40 & 0.51 & \textbf{0.63} 
\\
Automation & 0.44 & \textbf{0.49} & 0.48 
\\
Views on gender & 0.43 & 0.44 & \textbf{0.57} 
\\
Sexual harassment & 0.40 & 0.44 & \textbf{0.47} 
\\
Biomedical, food & 0.51 & 0.55 & \textbf{0.60} 
\\
Gender, Leadership & 0.50 & 0.45 & \textbf{0.59} 
\\
America in 2050 & 0.43 & 0.41 & \textbf{0.46} 
\\
Trust in science & 0.52 & 0.50 & \textbf{0.59} 
\\
Race & 0.38 & 0.42 & \textbf{0.51} 
\\
Misinformation & 0.48 & 0.48 & \textbf{0.54} 
\\
Privacy, Surveillance & 0.36 & 0.42 & \textbf{0.51} 
\\
Family, Relationships & 0.46 & 0.49 & \textbf{0.57} 
\\
Economic inequality & 0.38 & 0.47 & \textbf{0.55} 
\\
Global attitudes & 0.38 & 0.44 & \textbf{0.48} 
\\
Political views & 0.41 & 0.51 & \textbf{0.52}
\\
\bottomrule
\end{tabular}
\caption{Overall topic-wise accuracy based on exact match and collapsed match for answer choices. (demo.: demographic, ideo.: ideology, op.: opinions)}
\label{tab:topicwise-accuracy}
\end{table*}

\paragraph{Topic-wise Accuracy} 
Table \ref{tab:topicwise-accuracy} demonstrates the model's accuracy across different topics with various input sources, measured by exact match and collapsed match for answer choices. The model with demographic and implicit opinions particularly achieves higher scores on the Biomedical-food and Guns topics, implying that these two topics may lead the users to have similar opinions of each other. In contrast, the model exhibits slightly decreased performance when incorporating implicit opinions on topic Automation. This suggests that the LLM can make accurate predictions up to some extent based on user demographic and ideology information. However, incorporating implicit opinions, which may include viewpoints not aligned with users' demographic or ideologies, can potentially confuse the model in its prediction process.

\begin{figure}[!h]
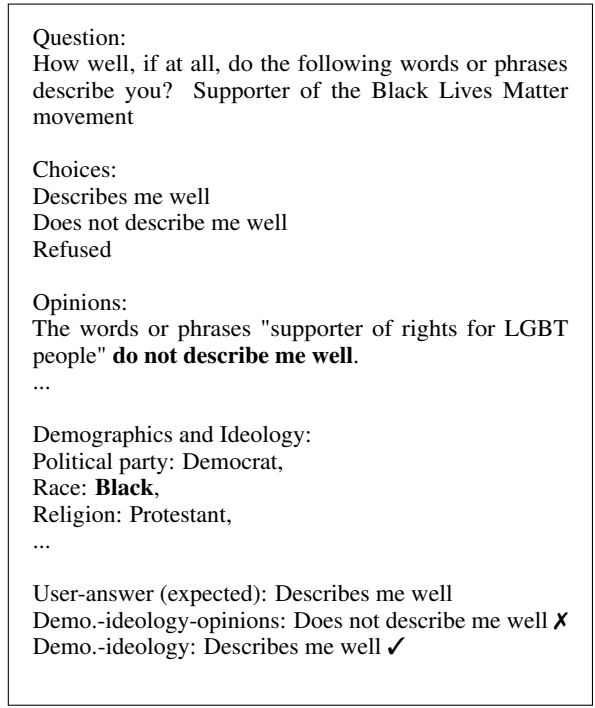

\small
\begin{framed}

Question: \\
How well, if at all, do the following words or phrases describe you? Supporter of the Black Lives Matter movement \\

Choices:\\
Describes me well\\
Does not describe me well\\
Refused \\

Opinions: \\
The words or phrases "supporter of rights for  LGBT people" \textbf{do not describe me well}. \\
...\\

Demographics and Ideology: \\
{Political party: Democrat}, \\
{Race: \textbf{Black}}, \\
{Religion: Protestant}, \\
...\\

User-answer (expected): Describes me well \\
Demo.-ideology-opinions: Does not describe me well \xmark \\
Demo.-ideology: Describes me well \cmark\\

\end{framed}
\caption{An example of a not relevant opinion confusing the model.}
\label{fig:error1}
\end{figure}

\paragraph{Common Errors} Figure \ref{fig:error1} is one of the most common errors when adding implicit opinions confuses the model. While the model makes a correct guess based on the person's demographic information for the question, after seeing implicit opinions having ``does not describe me well" that are also contained in the question's answer choices, the model got confused and makes an incorrect prediction.

\subsection{LLM with majority answer choices}
\label{sec:llm-majority}
Additionally, we also wanted to understand if similar performances can be achieved if we model an individual as a member of a (sub-)population, mirroring \cite{opinionqa}. 
For this, we first merge our QA data points using a particular ideological group value (e.g., democrat) and obtain the answer choice that is chosen by most of the group members (i.e., a majority vote) and treat that answer as the gold answer for the question while calculating the accuracy \cite{kim2023aiaugmented}. 

We prompt our model to predict an answer given a question assuming the role of a group representative, i.e., a person having a majority vote answers belonging to a specific group. The prompt that we used for this experiment can be found in Appendix \ref{appendix:prompt}. 

We see that the LLM is good at predicting the answer given by the majority of the group member belonging to a certain ideology, suggesting that LLMs are good at modeling a representative individual of a sub-population (e.g., all democrats). The overall performance without ideology information is 0.549 (with exact answer choice match) and 0.659 (with collapsed answer choice match), as presented in Table \ref{tab:performance-ideol}. This also indicated that the default opinions from the LLMs are somewhat aligned with the majority opinions seen at a population level. 

\begin{table}[!t]
\small
    \centering
    \begin{tabular}{lccc}
    \toprule
   & \bf Exact match & \bf Collapsed match \\ \midrule
Majority answer & 0.549  & 0.659 \\ \midrule
Independent & 0.546  & 0.674 \\ 
Democrat & 0.578  & 0.665 \\ 
Republican & 0.523  & 0.639 \\ \midrule
Avg overall & 0.566  & 0.667 \\ \bottomrule
    \end{tabular}
    \caption{Performance with LLM with ideology information.}
    \label{tab:performance-ideol}
\end{table}

\subsection{LLM as a person with an ideology}
\label{sec:llm-ideol}
We continue the same exercise, but we add the ideological information to see if this additional information can help the LLM perform better to model a user belonging to a group that believes in a specific ideology (e.g., conservative). The prompt that we used for this experiment can be found in Appendix \ref{appendix:prompt}. We find that the LLM is moderately good at modeling a user with group-level information to predict  the group-level majority opinion. This indicates that the additional ideological information is not particularly helpful. The overall performance with ideology information is 0.566 (with exact answer choice match) and 0.667 (with collapsed answer choice match), as shown in Table \ref{tab:performance-ideol}. We see a similar trend in results for modeling an individual with their demographics and/or ideology and/or past opinions since an individual's opinion does not align with the group's majority opinion that the person belongs to.

\section{Discussion and Conclusion} 
\label{sec:discussion}
An aligned LLM offers the benefit to offer personalized perspectives that align with a user's values, and cultural beliefs. However, there exist circumstances when LLMs can become an amplifier for unethical and biased views. 

\paragraph{Ethical concerns} With an aligned LLM, users can select information that adheres to their system of beliefs and to amplify potentially biased and unethical views. Such an echo chamber \cite{del2016echochambers} can eventually cause harm by reinforcing undesirable or polarized a user's views. 

A viable mitigation is to show user demography or ideology group answers in addition to the personalized answer (e.g., showing how an average Democrat with similar demographics would think on this topic and why). Further, past opinions can be used to ground an explanation (e.g., the current personalized answer is influenced by a user's specific past opinion), thus offering an opportunity for the user to introspect their past opinions.

\paragraph{Extensions}
Our work lays the foundation for a robust LLM alignment approach. By using memory-based personalization and recording interactions saved in a growing memory, the model can inform future instances of the \textit{most relevant} past opinions. Further, the interaction between demographics and opinions can be made seamless with a simulated annealing method that increasingly relies on user opinions as the memory grows and backs off to the group level/demographics-based opinion.

\paragraph{Conclusion}
This paper offers a new insight that aligning LLMs to users is best done by modeling user demographics, ideologies, and the most relevant past opinions. Large-scale experiments on PEW surveys present in the OpinionQA dataset show an approximately 7\% absolute QA accuracy over strong demography-based baselines. We proactively offer suggestions to avoid personalized LLMs from becoming echo chambers. An exciting future direction is to continuously store user opinions and grow the memory of opinions.

\section*{Acknowledgements}
We thank the members of the Aristo team at AI2 and Kurt Gray for their insightful feedback on this work. EH was funded, in part, by the Vector Institute for AI, Canada CIFAR AI Chairs program, an NSERC discovery grant, and a research gift from AI2. BPM was funded, in part, by an Adobe Research Fellowship.

\bibliography{anthology,custom}
\bibliographystyle{acl_natbib}

\appendix

\section{Prompt}
\label{appendix:prompt}
We provide a comprehensive display of all prompts used in the models incorporating user demographics, ideology, and opinions, which were employed for individual-user level tests in Figure \ref{fig:prompt-imp} and \ref{fig:prompt-exp}. Additionally, we present the prompts utilized for experiments conducted at the group-level tests in Figure \ref{fig:prompt-ideology} and \ref{fig:prompt-average}.

\begin{figure*}[!t]
\begin{framed}
A person has the following opinions on Guns. \\

Opinions: \\
1. The person used air guns, such as paintball, BB or pellet guns, sometimes when they were growing up. \\
2. The ease with which people can illegally obtain guns contributes a great deal to gun violence in the country today. \\
3. I attend gun shows sometimes. \\
4. I worry a little about having a personal health crisis. \\
5. People in my local community tend to look at most gun owners in a positive way. \\
6. Sport shooting, including target shooting and trap and skeet, was a reason why there were guns in my household when I was growing up. \\
7. The most important reason why I own a gun is for sport shooting, including target shooting and trap and skeet. \\
8. I sometimes visit websites about guns, hunting, or other shooting sports. \\

Based on the above list of opinions, which answer choice will this person select for the question: \\

Question: Thinking about gun owners who do not have children in their home how important do you think it is for them to: Take gun safety courses \\

Answer choices: \\
A.Essential \\
B.Important but not essential \\
C.Not important \\
D.Should not be done \\
E.Refused \\

Answer:
\end{framed}
\caption{prompt for implicit-only model}
\label{fig:prompt-imp}
\end{figure*}

\begin{figure*}[!t]
\begin{framed}
A person can be described as follows: \\

Age: 30-49 \\
Citizenship: Yes \\
Region: South \\
Education: Postgraduate \\
Income: $75,000-$100,000 \\
Marital status: Never been married \\
Political ideology: Conservative \\
Political party: Republican \\
Race: White \\
Religion: Roman Catholic \\
Frequency of religious attendance: A few times a year \\
Gender: Male \\

Based on the demographic information, which answer choice will this person select for the question: \\

Question: Thinking about gun owners who do not have children in their home how important do you think it is for them to: Take gun safety courses \\

Answer choices: \\
A.Essential \\
B.Important but not essential \\
C.Not important \\
D.Should not be done \\
E.Refused \\

Answer:
\end{framed}
\caption{prompt for demographic-only model}
\label{fig:prompt-exp}
\end{figure*}

\begin{figure*}[!t]
\begin{framed}
Thinking of yourself as a [republican/independent/democrat], please select the right choice.

{ques}

Choice: ["choice1", "choice2", "choice3"]

\end{framed}
\caption{prompt for ideology test}
\label{fig:prompt-ideology}
\end{figure*}

\begin{figure*}[!t]
\begin{framed}
Thinking of yourself as a person, please select the right choice.

{ques}

Choice: ["choice1", "choice2", "choice3"]
\end{framed}
\caption{prompt for LLM average person ability test}
\label{fig:prompt-average}
\end{figure*}

\section{
Similar ideologies and different opinions}
We show the percentage of user pairs having similar ideologies and the percentages of user pairs having similar opinions and different opinions within the user pairs sharing similar ideologies in Table \ref{tab:same_demo_diff_op}.

\begin{table*}[!t]
\small
    \centering
    \begin{tabular}{lccc}
    \toprule
  Topic & similar ideol. user pair (\%) & similar ideol. \& op. (\%) & similar ideol.-diff. op. (\%) \\ \midrule
Guns & 16.05  & 52.67 & 47.33 \\ \midrule
Automation & 15.99  & 15.26 & 84.74 \\ \midrule
Views on gender & 16.26  & 39.58 & 60.42 \\ \midrule
Sexual harassment & 16.95  & 22.14 & 77.86 \\ \midrule
Biomedical, food & 15.69  & 13.44 & 86.56 \\ \midrule
Gender, Leadership & 17.52  & 50.25 & 49.75 \\ \midrule
America in 2050 & 15.34  & 30.21 & 69.79 \\ \midrule
Trust in Science & 15.53  & 26.27 & 73.73 \\ \midrule
Race & 16.32  & 21.05 & 78.95 \\ \midrule
Misinformation & 16.22  & 34.83 & 65.17 \\ \midrule
Privacy, Surveillance & 16.13  & 22.14 & 77.86 \\ \midrule
Family, Relationships & 17.01  & 48.88 & 51.12 \\ \midrule
Economic inequality & 16.24  & 39.25 & 60.75 \\ \midrule
Global attitudes & 16.75  & 47.78 & 52.22 \\ \midrule
Political views & 16.65  & 37.59 & 62.41 \\ \bottomrule
    \end{tabular}
    \caption{Percentage of user pairs sharing similar ideologies (similar ideol. user pair) and the percentages of similar opinions (similar ideol. \& op.) and different opinions (similar ideol.-diff. op.) within user pairs sharing similar ideologies.}
    \label{tab:same_demo_diff_op}
\end{table*}

\end{document}

%% file: macros.tex
\usepackage{tikz}
\usepackage{pgfplots}
\usepackage{xspace}
\usepackage[frozencache,cachedir=.]{minted}

\definecolor{lightgray}{gray}{0.9}
\definecolor{Box1Color}{RGB}{227, 236, 246}
\definecolor{Box2Color}{RGB}{248, 220, 225}
\definecolor{Box3Color}{RGB}{255, 238, 224}
\definecolor{cbBlue}{RGB}{0, 114, 178}
\definecolor{cbOrange}{RGB}{240, 228, 66}
\definecolor{cbGreen}{RGB}{0, 158, 115}
\definecolor{cbRed}{RGB}{213, 94, 0}
\definecolor{cbPurple}{RGB}{204, 121, 167}
\definecolor{cbSkyBlue}{RGB}{86, 180, 233}
\definecolor{cbGray}{RGB}{128, 128, 128}
\definecolor{CBF1}{RGB}{255,99,132}  
\definecolor{CBF2}{RGB}{54,162,235}  
\definecolor{CBF3}{RGB}{255,206,86}  
\definecolor{CBF4}{RGB}{75,192,192}  
\definecolor{CBF5}{RGB}{153,102,255} 
\definecolor{CBF1b}{RGB}{205,89,112}  
\definecolor{CBF2b}{RGB}{44,142,215}  
\definecolor{CBF5b}{RGB}{133,92,225}  
\definecolor{PromptBgColor}{RGB}{255, 238, 224}

%% file: prompts/prompt-demo-ideo-opinion.tex
\begin{figure}[!t]

\begin{small}
    \captionsetup{justification=centering, labelfont=bf}
    
\begin{minted}[fontsize=\footnotesize, frame=lines, framesep=2mm, baselinestretch=1.2, breaklines, breaksymbolleft={}, breaksymbolright={},bgcolor=PromptBgColor]{text}
A person can be described as follows: 

Age: 30-49 
Income: $75,000-$100,000 
Political ideology: Conservative 
Political party: Republican 
Religion: Roman Catholic 
...

The person has the following opinions on Guns. 

Opinions: 
1. The most important reason why I own a gun is for sport shooting, including target shooting and trap and skeet. 
2. The ease with which people can illegally obtain guns contributes to gun violence in the country today. 
...

Based on the above list of opinions and the demographic information, which answer choice will this person select for the question: 

Question: Thinking about gun owners who do not have children in their home how important do you think it is for them to: Take gun safety courses 

Answer choices: 
A. Essential 
B. Important but not essential 
C. Not important 
D. Should not be done 

Answer:
\end{minted}

\end{small}
\caption{Prompt using demographics, ideology, and GPT embeddings based top-$k$ past opinions to predict the answer to a question.}
    \label{fig:prompt:imexp}
\end{figure}